\pdfoutput=1  

\documentclass[11pt]{article}

\usepackage[margin=1in]{geometry}
\usepackage{amsmath,amssymb,amsfonts}
\usepackage{algorithm}
\usepackage{algorithmic}
\usepackage{graphicx}
\usepackage{booktabs}
\usepackage{microtype}
\usepackage{xcolor}
\usepackage{authblk}
\usepackage{hyperref}
\usepackage{arydshln}
\hypersetup{
  colorlinks=true,
  linkcolor=blue,
  citecolor=blue,
  urlcolor=blue
}

\usepackage{bibentry}

\usepackage{subcaption}
\usepackage{amsthm}

\usepackage{amsmath}
\usepackage{amssymb}
\usepackage{mathtools}
\usepackage{bm}

\usepackage[numbers]{natbib}
\title{Aligning ESG Controversy Data with International Guidelines through Semi-Automatic Ontology Construction}

\author[1,2]{Tsuyoshi Iwata\thanks{\texttt{tsuyoshi.iwata@df.uzh.ch}}}

\affil[1]{University of Zürich (UZH),
  Plattenstr. 14, Zurich, 8032, Switzerland}

\affil[2]{RepRisk AG,
  Stampfenbachstrasse 42, Zurich, 8006, Switzerland}

\author[2]{Guillaume Comte}

\author[2]{Melissa Flores}

\author[3,4]{Ryoma Kondo\thanks{\texttt{kondor@g.ecc.u-tokyo.ac.jp}}}

\affil[3]{Graduate School of Information Science and Technology, The University of Tokyo, 7-3-1 Hongo, Bunkyo-ku, Tokyo, Japan}

\affil[4]{The Canon Institute for Global Studies, ShinMarunouchi Building 5-1 Marunouchi 1-chome, Chiyoda-ku, Tokyo, Japan}

\author[3,4]{Ryohei Hisano\thanks{\texttt{hisanor@g.ecc.u-tokyo.ac.jp}}}

\begin{document}

\maketitle

\begin{abstract}

The growing importance of environmental, social, and governance data in regulatory and investment contexts has increased the need for accurate, interpretable, and internationally aligned representations of non-financial risks, particularly those reported in unstructured news sources. However, aligning such controversy-related data with principle-based normative frameworks, such as the United Nations Global Compact or Sustainable Development Goals, presents significant challenges. These frameworks are typically expressed in abstract language, lack standardized taxonomies, and differ from the proprietary classification systems used by commercial data providers. In this paper, we present a semi-automatic method for constructing structured knowledge representations of environmental, social, and governance events reported in the news. Our approach uses lightweight ontology design, formal pattern modeling, and large language models to convert normative principles into reusable templates expressed in the Resource Description Framework. These templates are used  to extract relevant information from news content and populate a structured knowledge graph that links reported incidents to specific framework principles. The result is a scalable and transparent framework for identifying and interpreting non-compliance with international sustainability guidelines.

\end{abstract}

\section{Introduction}

Environmental, social, and governance (ESG) investing has become a global priority as financial institutions, regulators, and stakeholders increasingly integrate non-financial factors into decision-making processes~\citep{sherwoodResponsibleInvestingIntroduction2018}. Recent policy developments such as the European Union’s Corporate Sustainability Reporting Directive (CSRD)~\citep{eu_csrd_2022} and Sustainable Finance Disclosure Regulation~\citep{eu_sfdr_2019} represent emerging regulatory frameworks aimed at standardizing ESG disclosures. In parallel, global normative frameworks such as the United Nations Global Compact (UNGC)~\citep{un_ungc_2000} and Sustainable Development Goals (SDGs)~\citep{un_sdg_2015} promote broader sustainability principles and ethical commitments. Together, these developments reflect a broad movement toward more comprehensive and transparent ESG reporting. As a result, there is a growing demand for reliable ESG data that capture emerging risks, sector-specific impacts, and developments across supply chains and financial markets~\citep{EvolutionESGReporting2020}.

Meeting this demand remains a complex challenge. Traditional ESG data sources, including corporate sustainability and annual reports~\citep{osman2024knowledge}, are typically self-reported and often emphasize positive performance, which can introduce bias and limit coverage. Independent sources such as news media have therefore become essential for uncovering controversies, verifying corporate claims, and identifying emerging risks~\citep{angioniExploringEnvironmentalSocial2024}. ESG data derived from such news coverage are often referred to as ESG controversy data because they highlight adverse incidents that may signal underlying risks not captured in official disclosures~\citep{elamer2024esg}. However, news articles are written in free text, published in high volume, and distributed across a wide range of outlets and narrative threads. This complexity is further heightened by issues such as bias and misinformation. To manage this, commercial vendors annotate news using proprietary ESG classification systems. For example, MSCI applies 28 categories~\citep{msci_esg_research_msci_2024}, Sustainalytics uses 22~\citep{morningstar_morningstar_2024}, Refinitiv uses 24~\citep{berg_is_2021}, and RepRisk uses 108~\citep{reprisk_methodology_2}. These proprietary taxonomies differ significantly in structure and terminology. Because the design and application of these systems are often opaque, the resulting ESG ratings lack consistency across providers and are difficult to align with both formal regulatory frameworks and high-level normative frameworks~\citep{berg2022aggregate_confusion, billio_inside_2021}. This is particularly problematic for principle-based frameworks such as the UNGC, which are normative in nature and do not prescribe specific metrics~\citep{andrews_global_2020, frecautan_who_2022}. In practice, this makes it necessary to analyze original news content directly to ensure the accurate identification of ESG events and alignment with relevant standards.

Even with access to relevant ESG news content, aligning extracted information with ESG frameworks presents additional complexity. Normative frameworks such as the UNGC and SDGs, in addition to regulatory instruments such as the CSRD, are typically principle based and written in abstract, high-level language. These frameworks often lack standardized taxonomies and are subject to ongoing reinterpretation, amendment, or contextual adaptation. Mapping unstructured news content to such evolving frameworks requires more than static keyword tagging or rule-based systems. For example, assigning a general class such as ``human rights violations'' may overlook new distinctions introduced in revised guidelines, such as variations in due diligence expectations or cross-border responsibilities. These semantic mismatches contribute to the well-known discrepancies in ESG ratings among major data providers and limit the ability of stakeholders to draw consistent conclusions~\citep{berg2022aggregate_confusion, billio_inside_2021}.


\begin{figure}[t!]
\centering
\includegraphics[width=0.9\textwidth]{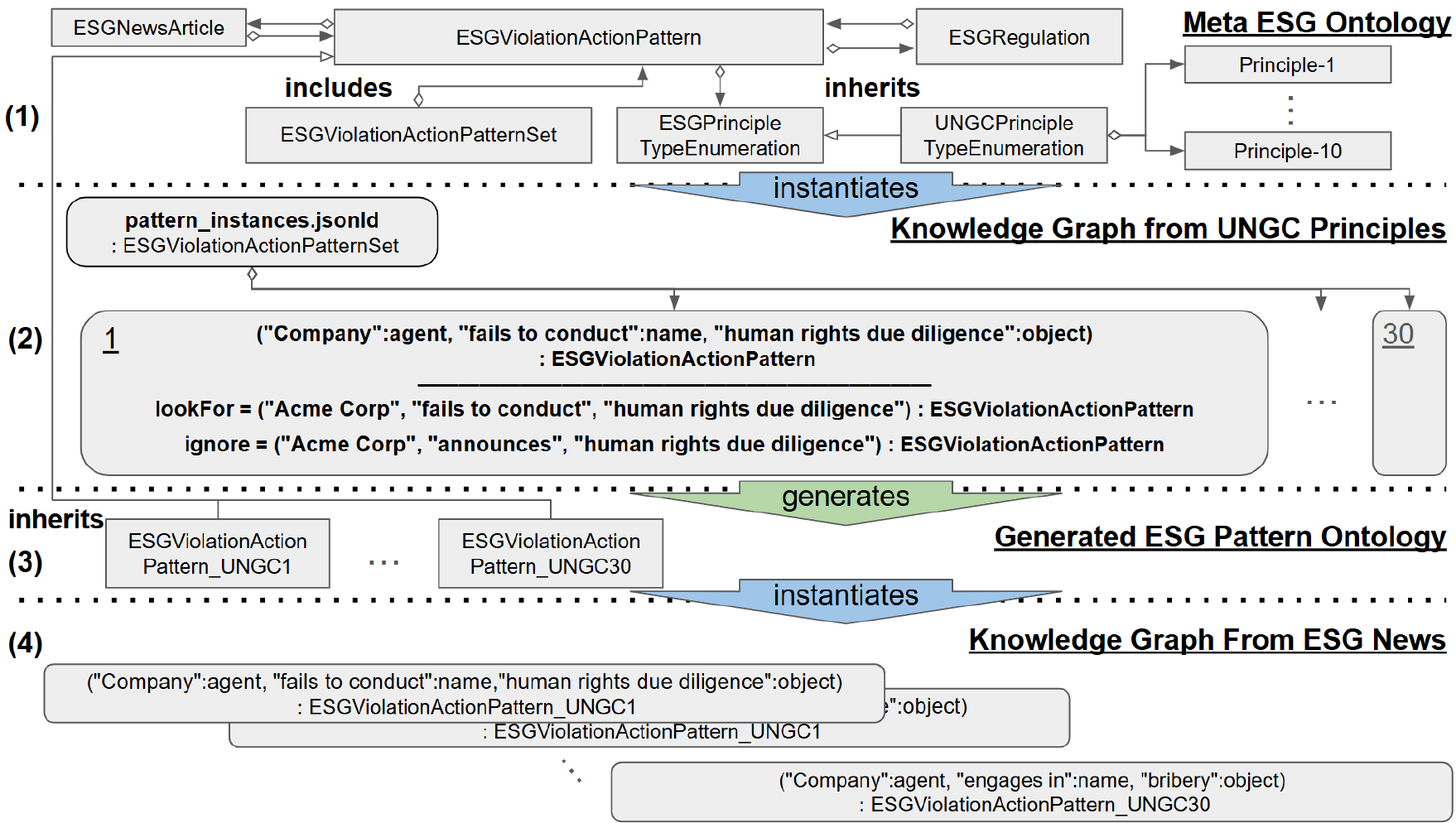}
\caption{Overview of the proposed method.}
\label{fig:figure1}
\end{figure}


To address the challenge of aligning unstructured ESG content with high-level normative frameworks, we propose a method for constructing ESG knowledge graphs using lightweight ontology design, pattern instantiation, and structured information extraction, all supported by large language models (LLMs). As illustrated in Figure~\ref{fig:figure1}, our approach begins with defining a meta-ontology that links ESG news, frameworks, and principles via violation patterns grounded in the UNGC framework (Figure~\ref{fig:figure1}(1)). For each principle, we use language models to generate concise Resource Description Framework (RDF) comments and three representative relational patterns, encoded as instances of an abstract violation pattern class (Figure~\ref{fig:figure1}(2)). Then we promote these patterns to 30 ESG violation templates (3 patterns \(\times\) 10 UNGC principles), each modeled as a dedicated RDF class to enable type-level reasoning (Figure~\ref{fig:figure1}(3)). Finally, we apply this framework to ESG-related news articles, where language models extract named entities and identify text segments that match the predefined patterns. We link matched triples to the relevant UNGC principles and instantiate them in a compliance-aligned knowledge graph (Figure~\ref{fig:figure1}(4)).

Our approach differs from those in previous studies on ESG knowledge graphs within the semantic web field because we align unstructured text directly with normative principles using automated, ontology-based techniques~\citep{angioniExploringEnvironmentalSocial2024,ongESGSenticNetNeurosymbolicKnowledge2025,osman2024knowledge,hisanoPredictionESGCompliance2020}. By matching extracted content to violation patterns associated with specific principles, we support interpretable detection of ESG non-compliance. These capabilities were not addressed in previous studies. Our approach is also connected to recent work on LLM-based ontology learning~\citep{BabaeiGiglou2023LLMs4OL}. We introduce ESG-aligned rdfs:Class definitions, which are integrated into existing ontologies through subclass inheritance, opening the door to a richer ESG knowledge graph. All code and data are publicly available: \url{https://github.com/tsuyoshiiwataRR/NEWS_LLM_UT_RR}.

\section{Methodology}

Figure~\ref{fig:figure1} presents our four-stage approach for constructing ESG knowledge graphs that align unstructured news content with formal ESG regulatory principles.

\textbf{(1) Meta ESG Ontology.}  
We begin by defining a lightweight ontology that connects ESG-related news, regulations, and principles through structured violation patterns. To achieve this, we extend \texttt{schema:NewsArticle} and \texttt{schema:Legislation} from Schema.org~\citep{schemaorg}, and \texttt{hna:ActionPattern} from our previous study~\citep{matsuoka2024hierarchical}, to introduce the classes \texttt{ESGNewsArticle}, \texttt{ESGRegulation}, and \texttt{ESGViolationActionPattern}. These classes are linked to ESG principles via \texttt{ESGPrincipleTypeEnumeration}, which we instantiate as \texttt{UNGCPrincipleTypeEnumeration} to represent the ten principles of the UNGC, covering human rights, labor, environment, and anti-corruption. Because of their abstract and normative phrasing, these principles are difficult to operationalize directly, which makes them the ideal benchmark for evaluating principled event extraction. To support this alignment, we use LLMs to generate concise \texttt{rdfs:comment} summaries for each principle. These serve as semantic anchors for downstream processing. We also define \texttt{ESGViolationActionPatternSet} to organize multiple patterns associated with each principle. All classes are aligned to upper ontologies (e.g., Schema.org) and typed with explicit domains/ranges, enabling RDFS/OWL reasoning over subclass hierarchies and properties.

\textbf{(2) Pattern Instantiation.}  

To operationalize these abstract principles, we first collected the official textual descriptions of each UNGC principle from the United Nations Global Compact website~\citep{un_ungc_2000}. We then prompt LLMs to generate representative violation patterns from these descriptions. For each principle summary, the model outputs three relational patterns in the form \texttt{(Entity A, Action, Entity B)}, which we design to capture both direct and indirect violations (e.g., \texttt{(Company, violates, rights of indigenous communities)}). We encode these patterns as instances of \texttt{ESGViolationActionPattern} in JSON-LD format. Each instance includes positive and negative examples, which we specify through the \texttt{lookFor} and \texttt{ignore} properties, respectively. This structured representation forms the foundation for fine-grained pattern matching and subsequent ontology construction.

\textbf{(3) Semi-Automatic Ontology Construction.} 
Based on the instantiated patterns, we semi-automatically define 30 \texttt{rdfs:Class} entities (corresponding to the total number of patterns, i.e., \(3 \times 10\) = 3 patterns for each of the 10 UNGC principles) by promoting each \texttt{ESGViolationActionPattern} instance to a subclass of \texttt{ESGViolationActionPattern}. These subclasses represent specific types of ESG violations and provide type-level constraints for annotating actions in ESG-related news. This intermediate ontology layer enables principled reasoning over ESG behavior and facilitates alignment with abstract regulatory norms.


\textbf{(4) Knowledge Graph Construction.}  
To populate the graph, we use a large collection of negatively framed news articles sourced from Webz.io~\citep{webzio} because such articles are more likely to report instances of corporate misconduct. We apply a two-stage filtering process using multiple LLMs (GPT-4o mini~\citep{openai_gpt4o_mini_2024}, GPT-4.1~\citep{openai_gpt4_1_2025}, and Claude 3.7~\citep{anthropic_claude_3_7_2025}). First, we retain only English-language articles and exclude those unrelated to corporate or ESG topics. Second, we use LLMs to further refine the corpus, selecting articles that mention a company and describe a plausibly ESG-relevant negative event. This results in a final candidate set of approximately 800 to 2,000 articles, which depends on the language model.


For each article, we apply a structured prompting strategy to extract named entities categorized into organizations, persons, and locations, while excluding non-informative content such as author bylines or boilerplate text. Then we evaluate the article against a predefined set of 30 violation patterns. For each matched pattern, we use a carefully designed prompt to instruct the language model to extract a corresponding triple as in ~\citep{matsuoka2024hierarchical}. We ground the resulting triple using the identified named entities and linked to the appropriate UNGC principle. This ensures that everything is based on textual evidence.

\section{Evaluation}

To assess the effectiveness of our ontology-informed extraction method, we conducted a manual evaluation using 200 human-labeled samples. We randomly selected these samples from the refined corpus produced by GPT-4o mini during the article selection stage. This choice was motivated by the observation that GPT-4o mini retained the largest set of candidate articles compared to other models, providing a broader and richer pool of candidate samples. In contrast, GPT-4.1 and Claude 3.7 exhibited significantly more conservative selection behavior, which limited their utility as sources to build a diverse and representative evaluation dataset. The samples represented matched triples aligned with the UNGC Principles. We evaluated the method using standard metrics: precision, accuracy, and recall.


Table~\ref{tab:accuracy_comparison} provides a comparison across language models. GPT-4o mini achieved the best performance across all evaluated metrics. Additionally, our ontology-guided method outperformed a simple one-shot approach, which attempted to assign principles directly based on their short descriptions and the full article text. Although the one-shot method achieved relatively high recall, this came at the cost of significantly lower precision and accuracy. The method frequently generated unsupported predictions, introducing noise and reducing reliability. These findings underscore the value of our structured method, particularly in settings where precision is more important than recall.

Table~\ref{tab:ungc_accuracy_summary_gpto4mini} presents the results for our structured prompting pipeline by GPT-4o mini. The model achieved consistently high accuracy across all ten UNGC Principles, with particularly strong performance in correctly identifying non-relevant content. This was reflected in the high number of true negatives, which contributed significantly to overall accuracy. Precision varied by principle, with the highest value of 0.86 observed for Principle 1. Principle 6 yielded the lowest precision, excluding Principle 9, which reflected differences in how clearly certain violations were expressed in the articles.


Although overall performance was strong, certain challenges remain. Precision was occasionally reduced due to broad interpretations of principle definitions. For example, general descriptions of poor working conditions may be classified as forced labor, thereby producing false positives for Principle 4. Recall performance also varied. Principles 1, 4, 6, and 8 achieved relatively low recall, often because of indirect or nuanced language that implies rather than states a violation. For instance, references to surveillance or safety concerns may suggest human rights issues, but are not always recognized as such. By contrast, Principles 2, 5, and 7 demonstrated the highest recall. Violations under Principle 5, such as child labor, were rare but clearly expressed, thereby making them easier to identify. Principle 7 appeared frequently because of its broader scope.



\begin{figure}[htbp]
    \centering
    \caption{Transition probability heatmap in percentage, From a violation in a Principle to another violation in another date by a same company. Total number of the pairs is 181.}
    \includegraphics[width=0.9\textwidth]{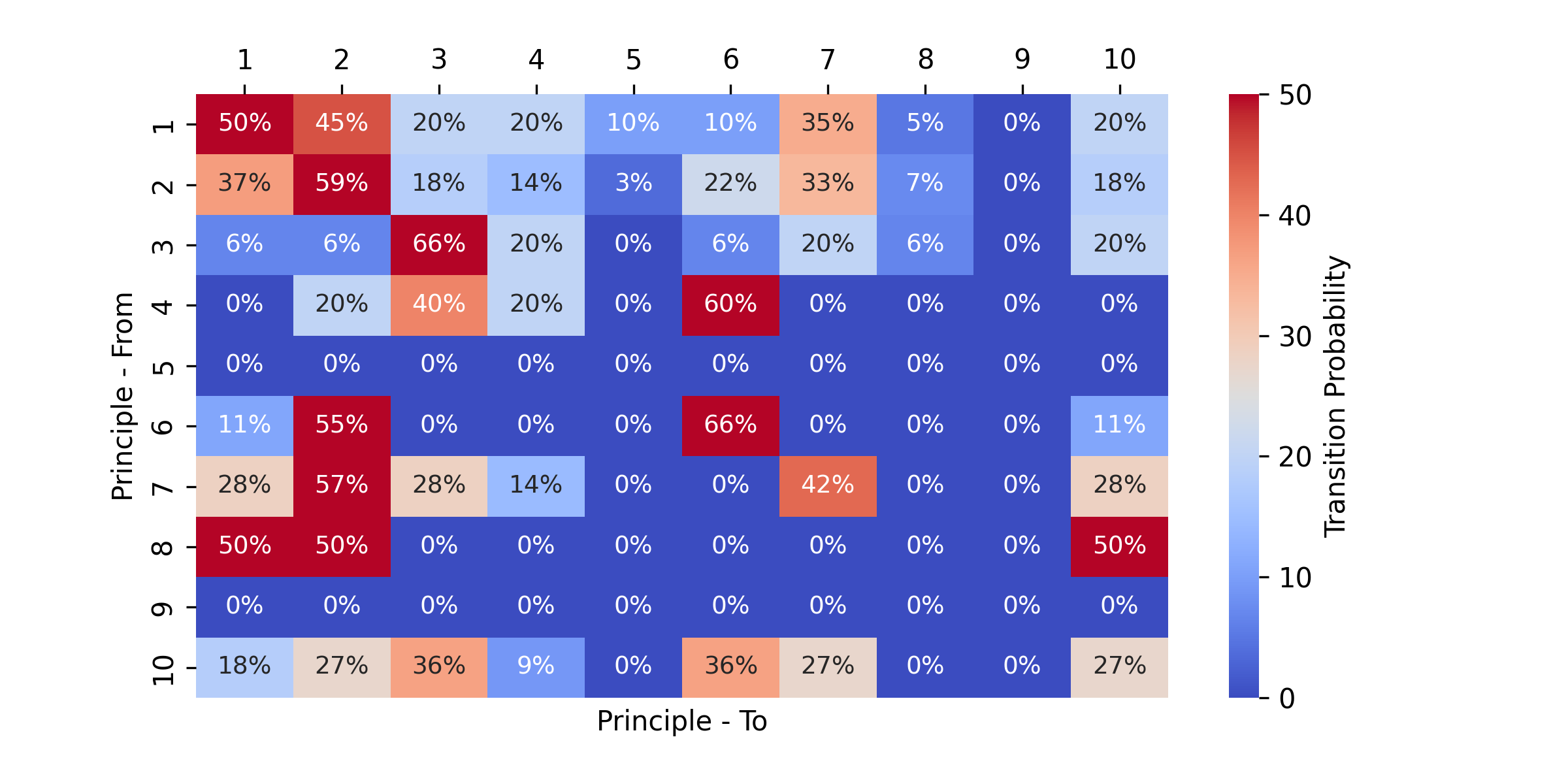}
    \label{fig:transition_heatmap}
\end{figure}

As an additional outcome of our experiments, we report the probability that a violation on one date is followed by another violation on a subsequent day by the same entity, as identified by our method, using the news dataset shown in Figure \ref{fig:transition_heatmap}. The figure indicates a recurring pattern of violations within the same Principle, as well as cross-Principle infringement patterns. However, a larger dataset would be needed to confirm the robustness of these findings.




\begin{table}
\caption{Comparison table among different LLMs.}
\label{tab:accuracy_comparison}
\centering
\label{tab:ungc_accuracy_summary}
\resizebox{0.8\textwidth}{!}{%
\begin{tabular}{lccccccc}
\toprule
Model & TP & FN & FP & TN & Accuracy & Precision & Recall \\
\midrule
GPT-4o mini & 80 & 116 & 62 & 1742 & 0.91 & 0.56 & 0.41 \\
GPT-4.1 & 31 & 165 & 24 & 1780 & 0.91 & 0.56 & 0.16 \\
Claude 3.7 Sonnet & 27 & 169 & 62 & 1742 & 0.88 & 0.30 & 0.14 \\
\hdashline
GPT-4o mini ('one-shot') & 166 & 30 & 411 & 1393 & 0.78 & 0.29 & 0.85 \\
\bottomrule
\end{tabular}}
\end{table}

\begin{table}

\caption{Summary of s Accuracy Metrics, among human-labeled 200 samples (GPT-4o mini)}
\label{tab:ungc_accuracy_summary_gpto4mini}
\resizebox{0.8\textwidth}{!}{%
\begin{tabular}{lccccccc}
\toprule
UNGC Principle & TP & FN & FP & TN & Accuracy & Precision & Recall \\
\midrule
 1 (Human Rights -Respect rights) & 25 & 41 & 4 & 130 & 0.78 & 0.86 & 0.38 \\
 2 (Human Rights -Avoid abuse)& 16 & 10 & 13 & 161 & 0.88 & 0.55 & 0.62 \\
 3 (Labour -Union rights)& 8 & 10 & 12 & 170 & 0.89 & 0.40 & 0.44 \\
 4 (Labour -No forced labour)& 4 & 10 & 3 & 183 & 0.94 & 0.57 & 0.29 \\
 5 (Labour -No child labour)& 4 & 3 & 2 & 191 & 0.98 & 0.67 & 0.57 \\
 6 (Labour -Equal opportunity)& 3 & 6 & 10 & 181 & 0.92 & 0.23 & 0.33 \\
 7 (Environment - Prevent harm)& 12 & 9 & 8 & 171 & 0.92 & 0.60 & 0.57 \\
 8 (Environment - Act sustainably)& 5 & 13 & 1 & 181 & 0.93 & 0.83 & 0.28 \\
 9 (Environment - Eco innovation) & 0 & 1 & 1 & 198 & 0.99 & 0 & 0 \\
 10 (Anti Corruption) & 3 & 13 & 8 & 176 & 0.90 & 0.27 & 0.19 \\
\hdashline 
All & 80 & 116 & 62 & 1742 & 0.91 & 0.56 & 0.41 \\
\bottomrule
\end{tabular}}
\end{table}

\section{Conclusion}

In this paper, we presented an ontology-informed method to align unstructured ESG news with high-level principles through pattern-based extraction and knowledge graph construction. Our approach integrates lightweight ontology design and LLM-driven pattern generation to produce semantically consistent RDF triples of potential ESG violations. Applied to the UNGC principles, it outperformed a baseline one-shot approach in both accuracy and interpretability. The resulting knowledge graph supports applications such as tracking ESG trends~\citep{angioniExploringEnvironmentalSocial2024} and predicting future risk events~\citep{hisanoPredictionESGCompliance2020}, which we leave for future work.

\section{Acknowledgments}
R.H. is supported by JST FOREST Program (JPMJFR216Q), JST PRESTO Program (JPMJPR2469), 
Grant-in-Aid for Scientific Research (KAKENHI, JP24K03043), 
and the UTEC-UTokyo FSI Research Grant Program. R.K. is funded by JST, ACT-X Grant
Number JPMJAX23CA, Japan. During the preparation of this work, the author(s) used ChatGPT, Grammarly in order to: Grammar and spelling check, Paraphrase and reword. After using this tool/service, the author(s) reviewed and edited the content as needed and take(s) full responsibility for the publication’s content. We also thank Edanz (https://jp.edanz.com/ac) for editing a draft of this manuscript.


\clearpage
\bibliographystyle{plainnat}
\bibliography{references_tsuyoshi_zotero,Hisano_RepRisk}

\end{document}